%% file: ISSRE-2.tex
\documentclass[conference]{IEEEtran}
\IEEEoverridecommandlockouts
\usepackage{cite}
\usepackage{amsmath,amssymb,amsfonts}
\usepackage{algorithmic}
\usepackage{graphicx}
\usepackage{textcomp}
\usepackage{xcolor}
\usepackage{multirow}
\def\BibTeX{{\rm B\kern-.05em{\sc i\kern-.025em b}\kern-.08em
		T\kern-.1667em\lower.7ex\hbox{E}\kern-.125emX}}

\usepackage{diagbox}

\usepackage{enumitem}
\usepackage[utf8]{inputenc}
\usepackage{xspace}
\usepackage{csquotes}
\usepackage{makecell}
\usepackage{arydshln}
\usepackage{url}
\usepackage{listings}
\usepackage{refcount}
\usepackage{glossaries}
\input{glossary}

\newcommand{\mytilde}{\raise.17ex\hbox{$\scriptstyle\mathtt{\sim}$}}

\newif\ifDEVEL
\DEVELtrue  

\ifDEVEL
\newcommand{\mv}[1]{\textcolor{red}{#1}}
\newcommand{\jrc}[1]{\textcolor{orange}{#1}}
\newcommand{\done}[1]{\textcolor{green}{#1}}
\newcommand{\remove}[1]{\textcolor{Maroon}{#1}}
\else
\newcommand{\mv}[1]{}
\newcommand{\jrc}[1]{}
\newcommand{\done}[1]{}
\newcommand{\remove}[1]{}
\fi

\usepackage{eso-pic}

\begin{document}
		\AddToShipoutPictureFG*{%
			\AtPageUpperLeft{%
				\raisebox{-1.0cm}{%
					\makebox[\paperwidth][c]{%
						\parbox{0.9\paperwidth}{%
							\centering
							\small\itshape
							Preprint version of a manuscript accepted for publication in the
							\textit{30th IEEE International Symposium on Software Reliability
								Engineering (ISSRE) 2019}, Berlin, Germany.
						}%
					}%
				}%
			}%
		}
		
		\bstctlcite{IEEEexample:BSTcontrol} 
		
		
		\newif\ifANONYMIZE
		
		\ifANONYMIZE
		\newcommand{\propheticus}{Predictive\xspace}
		
		\else
		\newcommand{\propheticus}{Propheticus\xspace}
		\author{\IEEEauthorblockN{João R. Campos, Marco Vieira, Ernesto Costa}
			\IEEEauthorblockA{DEI/CISUC, University of Coimbra,\
				Portugal\\
				jrcampos@dei.uc.pt, mvieira@dei.uc.pt, ernesto@dei.uc.pt}
		}
		\fi
		
		\title{\LARGE \propheticus: Machine Learning Framework for the Development of Predictive Models for Reliable and Secure Software 
			\ifANONYMIZE
			\\
			\vspace{.3cm}
			\large Tools and Artifacts (TAR) Paper
			\else
			\thanks{
				Work partially funded by H2020 project ATMOSPHERE grant 777154 and FCT grant SFRH/BD/140221/2018.}
			\fi
		}
		
		\maketitle
		\thispagestyle{plain}
		\pagestyle{plain}
		
		\begin{abstract}
			The growing complexity of software calls for innovative solutions that support the deployment of reliable and secure software. \gls{ML} has
			shown its applicability to various complex problems and is
			frequently used in the dependability domain, both for supporting
			systems design and verification activities. However, using \gls{ML} is complex and highly dependent on the problem in hand,
			increasing the probability of mistakes that compromise the results. In this paper, we introduce \propheticus, a \gls{ML} framework
			that can be used to create predictive models for reliable and secure software systems. \propheticus attempts to abstract the complexity of \gls{ML} whilst being easy to use and accommodating the needs of the
			users. To demonstrate its use, we present two case studies (vulnerability prediction and online
			failure prediction) that show how it can considerably ease and
			expedite a thorough \gls{ML} workflow.

		\end{abstract}
		
		
		
		
		\section{Introduction}
		Technology is part of our everyday life and, as a result, it is becoming more and more common to execute critical and sensitive tasks through software. Consequently, the dependability of such solutions becomes of utmost importance, as significant consequences may arise otherwise. However, the growing complexity and size of software render many of the traditional techniques impractical. For example, effective techniques such as code reviews, inspections, and testing, are known to not scale properly nor in an affordable manner \cite{Fagan}. Thus, products are often sent to production containing faults that may compromise the supported business processes. 
		
		Due to recent technological developments, \gls{ML} algorithms have shown their ability to adapt and extract knowledge in various complex problems, and modern computational power allows their use on large datasets. In the dependability research field, \gls{ML} is commonly used to complement traditional techniques (e.g. reliability prediction \cite{Alsina2018}, error detection \cite{Nie2018}).
		
		An adequate and thorough \gls{ML} approach is complex and problem-dependent. It requires a deep understanding of the problem and its data to choose the more adequate techniques and algorithms. Moreover, it is very common that the available datasets present certain characteristics that make them difficult to process (e.g. imbalanced data, high and low dimensionality) and that must be properly handled. Besides defining which techniques to use, there are many other concerns, such as how to estimate the performance of the models and how to properly compare different solutions. All these choices lead to the necessity of mastering both theory and practice, as implementing all the concepts and techniques is far from trivial. Furthermore, a single small mistake in the experiment can be enough to undermine the whole process. 
		
		As \gls{ML} became widely adopted, various platforms have been developed (e.g. Weka \cite{eibe2016weka}, H\textsubscript{2}O.ai \cite{h2o}) that abstract its technical details. However, most cannot be easily customized or extended, contain small or limited libraries, have small communities and consequently evolve slowly, or abstract on a lower level and still require significant coding. Consequently, there is no commonly accepted tool within the dependability community, where each researcher uses \gls{ML} differently, often limiting the extent of the experiment (e.g. both \cite{Eshete2017} and \cite{Sauvanaud2016} consider only a single algorithm, based on similar work, without considering any other techniques). In fact, albeit comprehensive tools such as Weka are often used, there are still many researchers that resort to lower-level libraries, such as Scikit-learn \cite{scikit} (e.g. \cite{Eshete2017}). This suggests that, although such tools are adequate for certain purposes, they are not flexible or easily adaptable for many others.
		
		This paper introduces \propheticus\footnote{\label{footnote:demo}Code and demo available at \url{http://www.joaorcampos.com/ISSRE-2019}}, a framework that includes functionalities for all the steps in a \gls{ML} work, from data analysis and preprocessing, to model assessment and comparison. It was created to overcome the limitations of existing tools towards research in the dependability area, which requires it to be flexible and adaptable to fit the needs of the users. \propheticus can be applied to a variety of problems (e.g. error detection, failure prediction, intrusion detection) to create  models whose predictions can then be used to develop and deploy more dependable systems. To use it on a given problem, the user only needs to launch its \gls{CLI} as a normal Python script. Then, he can navigate through the menus to explore his data and execute the various tasks. Finally, the user can analyze and compare the results of different approaches to determine the best solution. 
		
		Two cases studies are presented on how \propheticus can be used to assist with dependability works. The first regards software vulnerabilities and attempts to create models that can predict vulnerabilities in the code based on software metrics. The second case is concerned with residual faults and focuses on \gls{OFP}, a technique that attempts to create models that are able to predict failures in the near future using past data and the current system state. The case studies show that \propheticus can be used to assist in both problems, leveraging the capabilities of \gls{ML} and facilitating its use, thus allowing the user to focus on the problem itself. 
		
		The paper is organized as follows. \textit{Section \ref{sec:related}} presents background concepts and related work. \textit{Section \ref{sec:propheticus}} and \textit{\ref{sec:case_studies}} include a high-level overview of the \propheticus framework and two case studies, respectively. \textit{Section \ref{sec:demo}} provides a short demo of \propheticus. Finally, \textit{Section \ref{sec:conclusion}} presents a brief discussion on the framework's novelty and concludes the paper.

		\section{Related Work}
		\label{sec:related}
		
		This section provides a brief overview of \gls{ML} and existing tools, as well as the relation between dependability and \gls{ML}. 
		
		\subsection{\acrfull{ML} Concepts}
		The recent technological developments brought \gls{ML} (back) to the spotlight due to its ability to adapt and extract knowledge from complex problems. One of the main reasons is that it can be used to find intricate patterns in the data without relying on a predetermined model. However, properly using \gls{ML} requires expert knowledge and consists of several steps. Depending on the feedback available to the learning system, \gls{ML} tasks can be classified into different types (e.g. \textit{Supervised}, \textit{Unsupervised}) \cite{bishop2006pattern}, each with its own set of techniques.
		
		\textit{Data preparation} (e.g. \textit{transformation, feature selection}) deals with the fact that often the data is not perfect (e.g. noise, redundancy) and needs to be prepared \cite{Kotsiantis2006}. \textit{Data analysis}, which encompasses \textit{Descriptive} (statistical, e.g. \textit{mean, standard deviation}) and \textit{Exploratory} (graphical tools, e.g. \textit{boxplots}) analysis, is used to gain the necessary insight on the problem  to assist in choosing the most adequate techniques. 
		
		A set of algorithms must be selected considering the characteristics of the problem. However, each algorithm usually has various parameters that can take a wide range of values and require a \textit{fine-tuning}. To obtain a realistic estimate of the prediction error of a model, \gls{ML} usually deals with two main datasets: \textit{training} (to fit the model) and \textit{test} (to estimate the generalization error of the model) \cite{bishop2006pattern}. This division is not trivial, thus several techniques are available (e.g. \textit{partition}). There are also various metrics to measure the performance of a model (e.g. \textit{recall}), but they should be carefully chosen, as they are not independent of the data \cite{Sokolova2009}. Samples that are correctly predicted are known as \textit{\gls{TP}} and \textit{\gls{TN}}. The positive samples that are predicted as negatives are \textit{\gls{FN}} and the opposite are \textit{\gls{FP}}. Finally, \textit{statistical comparisons} (e.g. \textit{ANOVA}) are required to determine the best model.
		
		Besides knowing which techniques to use, it is necessary to know when to use them, otherwise, several issues may arise (e.g. data leakage) that compromise the results. Moreover, implementing the code for the experiments is not trivial. Between mastering theory and practice, even minor misinterpretations can lead to faults that can invalidate the experiments.

		\subsection{\acrfull{ML} Libraries and Tools}
		As \gls{ML} became widely used, various general-purpose tools have been developed at both open-source and enterprise contexts. For research purposes, enterprise solutions (e.g. Feedzai \cite{feedzai}, DataRobot \cite{datarobot}) are not accessible, however, amongst the open-source tools, there is also no straightforward choice. 
		
		One of such tools, H\textsubscript{2}O.ai \cite{h2o}, is a \gls{ML} platform that has been gathering support (mostly at enterprise level). It includes several \gls{ML} algorithms and techniques and a GUI, supported by several developers and a growing community. However, although it is currently open-source, it is owned by a small company, which can easily make it proprietary for profit. Additionally, its development practices are not very strict, resulting in hundreds of branches with failed tests in their Git repository, as well as some loose coding standards \cite{h2ogit}.
		
		Weka \cite{eibe2016weka} is probably one of the most well-known open-source \gls{ML} tools and it contains a wide array of algorithms and techniques for data mining tasks. It is mostly known for its GUI (although it also provides a CLI and API) which allows the user to experiment and visualize different techniques. However, concerning research, it presents some disadvantages, which is why it is often used on an exploratory basis. Although its GUI mostly removes the need to code, it is not intuitive and less common tasks are not easy to execute (e.g., combining multiple preprocessing techniques). Moreover, Weka is implemented in Java, which has a steep learning curve and often imposes a rigid and complex code structure. As a result, although it also provides an API, customizing and expanding it is not trivial. Weka was initially developed in academia and is mostly maintained by a small team (although there are also third-party packages). Thus, its code is often not clean or efficient and the development and adoption of state-of-the-art techniques can take some time. Combined with the fact that Weka does not have a very active community, support is rather poor, which is only worsened by its documentation. In fact, throughout the years, Weka has been losing relevance (as highlighted in \cite{Saez2017}) to more actively developed frameworks.
		
		One of the most promising solutions currently available is Scikit-learn \cite{scikit}, which is a comprehensive library of \gls{ML} algorithms and techniques. It has been widely adopted and thus has a large developer- and user-base. It is thoroughly documented which, combined with the active community, translates into good support for use and development. Its large number of developers allows it to steadily keep-up with advances in the state-of-the-art, all while ensuring peer-reviewing of its code towards standards and performance. It has an intuitive interface that is kept overall the framework, allowing it to seamlessly integrate with other relevant \gls{ML} packages (e.g. Imbalanced-learn \cite{imblearn}, Keras \cite{chollet2015keras}), and makes use of other scientific (e.g. numpy) and visualization packages (e.g. matplotlib). In fact, it has been gathering support even within the scientific community \cite{Eshete2017,Saez2017}. 
		However, it is made to be used only as an API and, although it provides several \enquote{helper} methods, it still requires significant coding for a sound/comprehensive experimental process. Besides the programming effort, this requires that the user has a considerable \gls{ML} knowledge to properly conduct the experiments according to theory. 
		
		\propheticus attempts to overcome most of these limitations by providing a generalizable, easy-to-use, and configurable structure that allows the user to take advantage of the developments of Python open-source \gls{ML} libraries. Relaying the implementation of \gls{ML} techniques to such libraries provides confidence in their results, as most of them are stable, peer-reviewed, and thoroughly tested. Additionally, the easy integration of new libraries and techniques allows the users to explore different approaches with minimal coding, thus reducing the probability of errors. Still, as it is written in Python, other functionalities are also fairly easy to implement.
		

		

		\subsection{Dependability and \acrfull{ML}}
		Dependability is a broad concept that encompasses reliability, availability, integrity, amongst other attributes \cite{AlgirdasAvizienisLaprieJean}. The threats to dependability are faults, which can manifest as errors, that can eventually lead to failures. Throughout the years, several techniques have been designed to increase dependability. These can be mainly divided into two large groups: \textit{fault avoidance} (aims for fault-free systems) and \textit{fault acceptance} (accepts and deals with the existence of faults). 
		
		Due to the growing complexity and sheer dimension of software, traditional techniques such as code reviews and testing are not able to scale well nor in an affordable manner \cite{Fagan}. Following the recent trend of using \gls{ML} in complex problems, there has been some research on applying it to the dependability domain, as \gls{ML} techniques can extract knowledge that most likely would otherwise not be found. In \cite{Alsina2018} the authors used \gls{ML} algorithms to predict the reliability of components and concluded that some achieve better results than the equivalent traditional technique. In \cite{Nie2018} the authors successfully used \gls{ML} to predict GPU errors in \gls{HPC} systems. Finally, in \cite{Alves2016a} the authors use a large vulnerability dataset and concluded that some \gls{ML} algorithms were able to predict all vulnerabilities using software metrics. 

		Although \gls{ML} has been recurrently used in the dependability domain, most works are both limited to a small set of methods and are specifically implemented according to the scope of the experiment. This reduces the ability to compare results and even minor decisions or bugs can drastically influence the resulting models. \propheticus attempts to address this issue by abstracting implementation details that are prone to error. Additionally, it also reduces the overhead of using \gls{ML} for a specific research purpose by providing a flexible framework that can be adapted to fit the requirements of the users.

		\section{\propheticus Framework}	
		\label{sec:propheticus}
		\propheticus emerged with the purpose of easing, automating, and assuring the workflow of a \gls{ML} approach applied to the dependability domain, but can also be customized to fit unforeseen uses. It provides a data-centric approach so the user can focus on the problem instead of the implementation. \propheticus does not intend to replace (or be better than) other alternatives, but rather it was developed based on the unfulfilled necessity of a tool focused on research in the dependability community.
		
		Although \propheticus does not attempt to remove the complexity of a \gls{ML} work, some of its process and rules are fairly standard. As a result, one of the main goals of the framework is that it should clearly define the workflow of the experiment, validating and providing useful feedback to the user along the process. Additionally, it also attempts to identify common mistakes which can usually be detected when they are known (e.g. data leakage), yet if not attended to will pass undetected and compromise the validity of the experiments. 
		
		\subsection{Overall Architecture}
		\label{sec:framework:arch}
\begin{figure*}[!t]
			\centerline{\includegraphics[width=\textwidth]{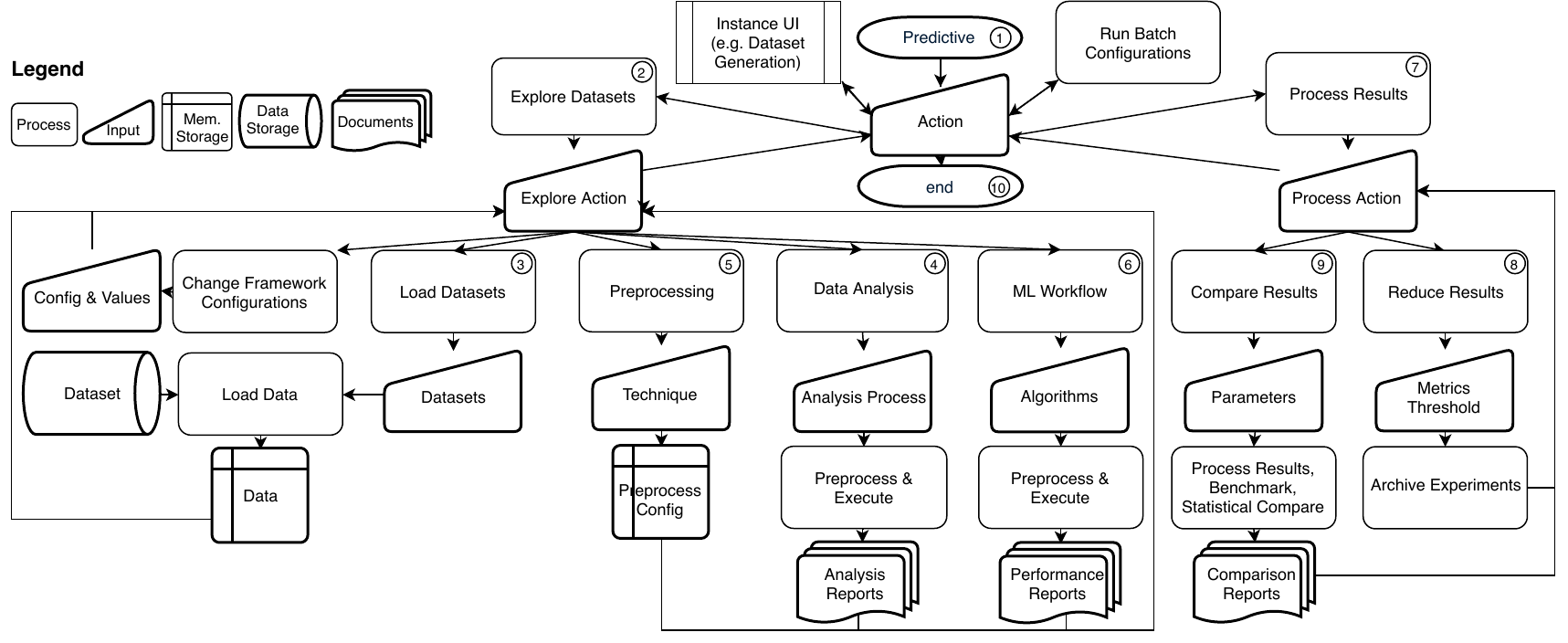}}
			\caption{\propheticus High-Level Architecture / Workflow}
			\label{fig:architecture}
		\end{figure*}

		A high-level overview of the architecture of \propheticus can be seen in \textit{Fig. \ref{fig:architecture}}. It is based on object-oriented programming, hence the different processes are encapsulated in classes. The modules are briefly described in \textit{Section \ref{sec:framework:modules}}.\\ 
		\indent At the moment, the use of \propheticus is based on a simple, yet comprehensive, \gls{CLI} that allows the user to intuitively explore and flow through the different steps. To accommodate user-specific configurations and code, the framework must be instantiated (i.e. specific folders and files must be created to define the problem scope and configurations) for each problem. Additionally, \propheticus contains hooks that allow the user to integrate functionalities (e.g. create datasets) with the \gls{CLI}.
		
		The framework expects the inputs datasets in a specific structure, comprised of two files. The data file follows a straightforward structure: a simple table where the columns are divided into a set of features and targets, and where each row represents a sample that contains a value for each feature and its targets. The second file is the headers file, which must contain a JSON object identifying the details of the features.
		
		For most functionalities, \propheticus creates reports that can later be used for analysis. These reports are stored on the hard drive and their filenames are hashed using all the experiment's configurations. This allows generating the same report for different configurations without conflict.
		
		\subsection{Implementation}
		\propheticus is implemented in Python, a programming language that is flexible, easy to use, and transversal to several research areas. Additionally, some of the most well-known \gls{ML} libraries are implemented in Python (e.g. Scikit-learn \cite{scikit}, Tensorflow \cite{tensorflow}). \propheticus leverages the research knowledge of open-source communities, and as such strongly relies on Scikit-learn for most of its \gls{ML} tasks (e.g. algorithms). Scikit-learn includes several \gls{ML} methods, is thoroughly documented, widely adopted and its community is quite large and active, resulting in regular updates that keep it up to date with stable state-of-the-art developments. 
		Other \gls{ML} sources such as Imbalanced-learn \cite{imblearn} (e.g. sampling), PyClustering \cite{pyclustering} (e.g. clustering), and SciPy \cite{scipy} (e.g. statistics) are also used to complement the functionalities of the framework.
		
		\propheticus runs on any system that supports Python. Currently, all processes are handled in-memory, which requires the machine to have enough memory to hold the data. However, other data management approaches will be considered in the future to improve scalability (e.g. online learning). 

		\subsection{Functionalities / Modules}
		\label{sec:framework:modules}
		\propheticus acts as a standalone application and its workflow is centered on the interface module that controls all process logic and calls the different modules. Although \propheticus is intended to be used iteratively, an \enquote{initial} use would follow the numeric order identified in \textit{Fig. \ref{fig:architecture}}.
		
		The \texttt{Data Management} module handles the process of loading and preparing the datasets. It allows the combination of multiple datasets (e.g. different sources) as long as they share the same structure. Each dataset must have a headers file, containing the details (e.g. name, type) of each feature. Additionally, it is also possible to define other details (e.g. if a variable is categorical, which will be automatically encoded) or domain knowledge (not yet implemented) that may be used to improve the process (e.g. how to handle missing values). 
		
		The \texttt{Data Analysis} module encompasses the logic associated with exploring and analyzing the datasets. It contains functionalities for both \textit{Descriptive} and \textit{Exploratory} analysis. 
		Another module, \texttt{Data Preprocessing}, includes the logic required for preprocessing the data. It contains functionalities to select/exclude only a subset of features, define which feature to use as \textit{target}, or select only samples with certain values for given features. This module also includes the logic for dimensionality reduction and data sampling.
		
		Probably the most important module, \texttt{Classification}, encompasses the execution of the whole experiment. The data is preprocessed according to the configurations and then passed to the selected \gls{ML} algorithms. By default, it executes the same configuration 30 times under different random seeds (a number commonly accepted as adequate under the \textit{Central Limit Theorem} \cite{Hogg2009}) and performs 10-fold \cite{Borra2010} stratified cross-validation (both configurations can be easily changed through the \gls{CLI}). Several metrics are considered to assess the performance of the models (e.g. $F_1$-score, informedness). These are computed fold-, run-, and experiment-wise. Upon finishing all the runs, the confusion matrix, the \gls{ROC} curve, and the Precision-Recall curve are generated, alongside a spreadsheet containing all the metrics and logs, which can then be used for further comparisons or analyses. As most algorithms contain various hyperparameters which can take a plethora of values, \propheticus allows fine-tuning them through grid-search. For this process, it uses a nested cross-validation approach (e.g. inner cross-validation is used to choose the parameters based on the training data) \cite{Cawley2010}. 
		A similar, but smaller, module, \texttt{Clustering}, focuses on clustering algorithms. Seemingly, various clustering metrics (e.g. silhouette) and reports are stored on a spreadsheet. 
		
		After getting some insight about the problem, users frequently want to conduct several experiments in an exploratory manner (e.g. various algorithms, datasets). As executing all the combinations manually is not practical nor scalable, \propheticus allows defining a list of configurations to execute in batch, which is used by the \texttt{Batch Execution} module. 
		
		When all the experiments are finished it is necessary to compare the results and identify the solution that best fits the needs of the users, which is implemented by the \textit{Process Results} module. One of its functionalities, \textit{Reduce Results}, allows the user to (re)move (move to a separate folder) experiments that have performances for certain metrics under given thresholds (e.g. move all experiments with 0 recall). However, the more prominent functionality within this module, \textit{Compare Results}, allows the user to choose the experiments that match given configuration parameters. This process generates a spreadsheet and various complementary graphs comparing the results (e.g. metrics, time complexity). \propheticus also explores the notion of application scenarios (a realistic situation of the problem in hand that depends on the criticality of the system) \cite{AntunesV15}, which allows the user to \textit{compare} and \textit{rank} models based on a specific set of metrics. Finally, this module also conducts \textit{statistical comparisons} between the experiments, informing the user if they are in fact different for a given significance level. 
		
		
		\subsection{Instantiation and Configuration}
		\propheticus aims at being generalizable and configurable. This is achieved by having client-specific structures that allow the user to configure the framework to his needs while minimizing the need to alter the framework's code. Moreover, it allows the user to have multiple projects under a single \propheticus installation. The customization of the framework is done mainly through three files: \texttt{InstanceConfig}, \texttt{InstanceGUI}, and \texttt{InstanceBatchConfiguration}.
		
		For a basic use (e.g. binary classification), it is not necessary to define any configurations. Still, framework- and problem-specific configurations (e.g. datasets location, binary/multi-class) can be specified in the \texttt{InstanceConfig} file.
		
		
		The \texttt{InstanceGUI} file is not mandatory, but if it exists \propheticus will add a custom menu at the first level of the \gls{CLI} to allow calls to user-specific menus (e.g. export datasets). Finally, the \texttt{InstanceBatchConfiguration} file can be used to generate/store a list of configurations of experiments that will be used by the \texttt{Batch Execution} module. 
		
		It is also possible to add new algorithms or techniques (e.g. dimensionality reduction) without modifying the code of the framework. This can be done by adding the new call details (e.g. package, method) to the existing list of the respective available methods. To integrate techniques that do not follow the Scikit-learn structure, a wrapper extending the required corresponding interfaces can be created in a folder specifically for that purpose (e.g. \texttt{algorithms}). If further, not yet customizable, changes are necessary (e.g. custom performance metrics), the structure of the framework is easy to understand and modify. Nonetheless, \propheticus will steadily evolve to allow configuring as much as possible.

		\section{Case Studies}
		\label{sec:case_studies}
		This section presents two case studies that demonstrate how \propheticus can be used in the dependability context. The first case focuses on vulnerability prediction (a \enquote{standard} classification problem) and conducts an exploratory analysis, uses different algorithms and techniques, and performs model selection by fine-tuning the hyperparameters of the algorithms. The second case is on \gls{OFP} (a time-series problem), where \propheticus is used to assess, rank, and statistically compare various models according to different criticality scenarios.

		\subsection{Vulnerability Prediction}
		Vulnerabilities are internal faults that allow external users to exploit and take advantage of the system. Software metrics have long been used to characterize software \cite{Shin2011,Khomh2012} and link them with existing vulnerabilities. Vulnerability prediction attempts to predict vulnerabilities in the code based on such software metrics. These predictions can then be used to guide the developers to remove them, thus improving dependability attributes such as reliability. Various works already use \gls{ML} for vulnerability prediction (e.g. \cite{Alves2016a}).
		
		This case study attempts to demonstrate how \propheticus can be used to explore such a problem and guide the search for the best techniques and algorithms. The evaluation uses a vulnerability dataset containing software metrics of functions, classes, and files pertaining to five different projects \cite{Alves2016}. In this work, we focus only on the data pertaining to the files of the Mozilla project as it is the largest and with more vulnerabilities. The resulting dataset pertains to 604304 files (out of which only 2819 are vulnerable) and contains 56 software metrics computed for each file.
		
		\subsubsection{Methodology}
		Due to the large dimension of the dataset and the significant data imbalance (as well as time and resource constraints), the dataset was limited to 100000 samples (still remaining a significantly imbalanced dataset). Initially, both a descriptive and an exploratory analysis were conducted. By observing the results from those analyses it was possible to perceive that most of the features' means were noticeably different between classes, suggesting they could be distinguishable. Additionally, by using heat maps and scatter plot matrices it was also possible to observe that there were some features with correlations above 90\%.
		
		Several techniques for dimensionality reduction, data sampling, and algorithms were used. The complete list can be seen in \textit{Table \ref{tb:vulnerabilities_experiments_configurations}}. For an initial approach, the hyperparameters of the algorithms were chosen based on existing literature and ad-hoc optimizations. Due to space restrictions, only the most relevant configurations are described next.
		
		Concerning dimensionality reduction techniques, \textit{Variance} removed features with 0 variance, whilst \textit{Correlation} removed highly correlated features (\textgreater 90\%). \textit{\gls{PCA}} kept the components that represent 99\% of the variance, and finally, \textit{\gls{RFE}} used the \textit{\gls{DT}} algorithm guided by the recall metric. Over- and undersampling methods were run with different ratios in relation to the number of samples in the minority class. Additionally, a Z-score normalization was used.
		
		Regarding the algorithms, different kernels (i.e. RBF, Linear, and Polynomial) were used for \textit{\gls{SVM}}, along with a relative gamma ($\frac{1} {n\_features}$) and penalty 1. The \textit{\gls{NN}} model was created with 100 neurons in a single hidden layer, the rectified linear unit activation function, and a stochastic gradient descent solver. The \textit{\gls{DT}} models used the Gini Impurity to decide the node splits with the \gls{CART} algorithm. \textit{Gradient Boosting} used a logistic regression loss function, and, same as \textit{Bagging, Extra Trees,} and \textit{\gls{RF},} used 100 estimators. To validate the results, 5-fold (instead of the default value 10, as the experiment also considers cross-validated fine-tuning) stratified cross-validation was used, and each experiment executed 30 times. 
		
		\begin{table}[!tb]
			\renewcommand{\arraystretch}{1.3}
			
			\caption{Experiments Configurations}
			\label{tb:vulnerabilities_experiments_configurations}
			\centering
			\begin{tabular}{r|l}
				\textbf{Parameter} & \textbf{Values}  \\ \hline
				Dim. Reduction & Variance, Correlation, \gls{PCA}, \gls{RFE} \\ 
				Sampling &\begin{tabular}{@{}c@{}}  
					Random under-/oversampling, SMOTE, Near Miss, \\ 
					ADASYN, Instance Hardness Threshold \end{tabular} \\
				Sampling Ratios & Oversampling: [1, 3] , Undersampling: [1, 4] \\
				Algorithms &   \begin{tabular}{@{}c@{}}  
					\gls{SVM}, Gradient Boosting, Bagging, \\  \gls{DT}, Adaboost, Extra Trees, \gls{NN},  \\ \gls{RF}, \gls{KNN}\end{tabular} \\ 
			\end{tabular}
		\end{table}
		
		An initial analysis was conducted using all the combinations of the different techniques in \textit{Table \ref{tb:vulnerabilities_experiments_configurations}} to identify which showed potential. As the hyperparameters of an algorithm can strongly influence its performance, a cross-validated fine-tuning through grid-search of each of the promising configurations was conducted. The metric used for fine-tuning was informedness, which measures how consistently a model predicts the outcome of both a \gls{TP} and a \gls{TN}.

		\subsubsection{Results}
		
\begin{figure}[!tb]
			\begin{minipage}[t]{0.48\linewidth}
				\begin{center}
					\includegraphics[width=1\textwidth]{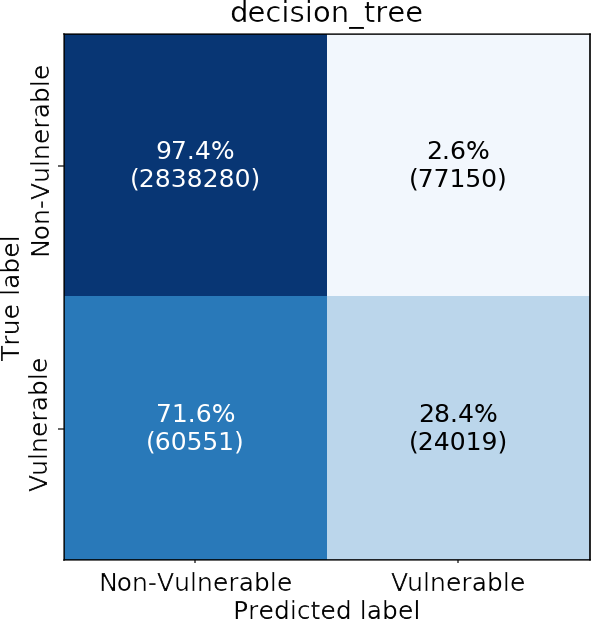}
				\end{center}
				\caption{Initial \gls{DT}}
				\label{fig:step1_dt_cm}
			\end{minipage}
			\hspace{0.02\linewidth} 
			\begin{minipage}[t]{0.48\linewidth}
				\begin{center}
					\includegraphics[width=1\textwidth]{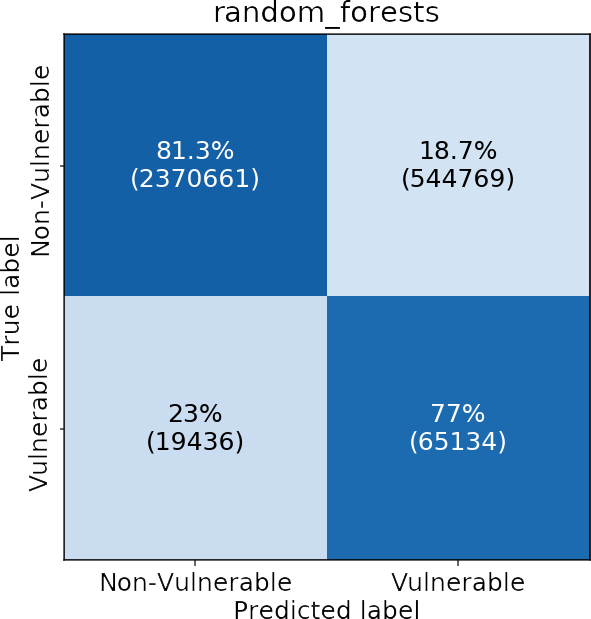}
				\end{center}
				\caption{\gls{RF} Hard. Thr., Var. \& Corr.}
				\label{fig:step5_rf_cm}
			\end{minipage}
		\end{figure}

		When using the algorithms on the entire dataset without preprocessing none of them was able to acceptably distinguish between the non-vulnerable and vulnerable samples. \gls{DT} produced the best models, yet, it was only able to predict 28\% of the vulnerable samples, as can be seen in \textit{Fig. \ref{fig:step1_dt_cm}}. These are fairly useless models as, although they do not have many \glspl{FP}, they cannot predict the vulnerable samples.
		
		As it is well known that imbalanced data can influence the performance of the algorithms \cite{He2009}, we initially used \textit{Random Undersampling}, which  lead to a significant improvement on the ability of the models to predict vulnerable samples (e.g. both \gls{RF} and Bagging were able to correctly predict 75\% of non-vulnerable and 82\% of vulnerable samples). However, the number of \glspl{FP} rose considerably, rendering the precision of the models very low. Nonetheless, these are still realistic models as there is a good relationship between the percentage of \glspl{TN} and \glspl{TP}. By combining the undersampling approach with dimensionality reduction it allowed some algorithms to improve, however, there were no significant changes. 
		
		As the resulting models were not yet satisfactory, the following approach was to combine both over- and undersampling methods (in this order). Initially, the method \textit{Random Oversampling} (a simple but usually effective method) was used, yet it was swiftly changed to \textit{\gls{SMOTE}} due to poor results. Combining it with dimensionality reduction lead to some more interesting results: \gls{RF} with feature selection by correlation was able to correctly predict 88\% and 69\% of non-vulnerable and vulnerable samples respectively. Although there is some loss of \glspl{TP}, the improvement in predicting non-vulnerable samples meant that there were significantly fewer \glspl{FP} (i.e. for 100k samples, the 13\% improvement corresponds to \mytilde13k less \glspl{FP}). As the results suggested that the combination of under- and oversampling techniques could lead to better models, several other sampling methods were used (e.g. \textit{\gls{ADASYN}}). The use of \textit{Instance Hardness Threshold} generated the most promising models for various algorithms. 
		
		Finally, a fine-tuning of the hyperparameters of the algorithms was conducted for the methods that showed potential. Overall, \gls{RF} with sampling by Instance Hardness Threshold, and feature selection by variance and correlation presented the best results, predicting 81\% of non-vulnerable and 77\% of vulnerable samples (as seen in \textit{Fig. \ref{fig:step5_rf_cm}}). However, as the "best" model depends on the needs of the user, using another metric in the fine-tuning process would create different solutions.	
		\subsubsection{Discussion}
		Due to the nature of the data, most algorithms struggled in predicting the vulnerable samples. By combining them with different techniques some improved their performance, although at the expense of having more \glspl{FP}. However, through the use of fine-tuning, it was possible to create models that could correctly predict most of the non-vulnerable and vulnerable samples. It would also be interesting to conduct a clustering study to check for groups or patterns of samples which may give further insight. Such a study could also be done using the \propheticus clustering module.
		
		This case study briefly demonstrated how \propheticus can be used for a complete \gls{ML} approach, from data analysis and preprocessing, to the execution of the models and performance assessment. By using \propheticus it is possible to easily configure, experiment, and combine the different techniques and assess their performance with minimal code modifications. 
		
		As an example, if we were using only Scikit-learn in this work, that would require significant effort (besides assuming the user possesses the necessary knowledge). All utilitarian functionalities (e.g. filter features), analyses (e.g. boxplots) and reports (e.g. confusion matrices) would have to be implemented. Also, the integration, execution, and assessment of the various techniques would have to be done individually as per the theory. 
		Although there are tools with graphical interfaces (e.g. Weka), they are not easy to customize or expand (e.g. to include more algorithms), inherently constraining their use.	
		\subsection{\acrfull{OFP}}
		Faults that make it to deployment can eventually lead to failures at runtime. \gls{OFP} intends to mitigate the effects of such residual faults, by using past data and the current system state for predicting the potential occurrence of failures \cite{Salfner2010}. This allows taking preemptive measures to avoid such failures or lessen their consequences, thus improving dependability attributes such as availability and reliability. Various approaches relying on \gls{ML} have already been proposed for \gls{OFP}\cite{Salfner2010}.
		
		The prediction of failures is based on different kinds of information, including the past data from the system, the current state of the system, the time horizon of the prediction, among others. As defined by Salfner and Malek in \cite{Salfner2010}, a prediction performed at time $t$ targets a window starting at time $t+\Delta t_l$, and lasting for $\Delta t_p$ ($\Delta t_l$ and $\Delta t_p$ are normally referred as lead-time and prediction-period, respectively). 
		
		Selecting a predictive model requires a rigorous assessment of alternative solutions using adequate metrics. Besides what was shown in the previous case study (in fact, \propheticus has already been used for a similar approach for \gls{OFP}\cite{Camposa}), \propheticus can also be used to assist in comparing and ranking models according to the preferences of the user. By exploring the notion of application scenarios, \propheticus  allows a better match of the outcomes of the models with the environmental requirements for the predictor operation. This case study considers four representative real-world scenarios with metrics that represent their scope (adapted from Antunes et al. \cite{AntunesV15}): \textit{business-critical} (i.e. every failure missed is problematic, but too many false alarms is not acceptable) which uses the $F_2$\textit{-score} metric, \textit{heightened-critical} (i.e. a few failures may be missed to lower false alarms) that uses \textit{informedness}, \textit{best-effort} (i.e. report few false alarms at the cost of missing some failures) that uses $F$\textit{-measure}, and \textit{minimum-effort} (i.e every false alarm raises concern, minimize them at the expense of missing failures) which uses \textit{markedness}. 
		
		This case study uses \propheticus to conduct an assessment and comparison of a large number of failure prediction models, including different criticality scenarios for choosing adequate metrics. The evaluation uses a dataset with failure data (generated through fault-injection) from Windows XP (SP3) \cite{Irrera2014}, comprising virtualized and non-virtualized environments. However, due to the scope of this case study, only the data of the non-virtualized environment will be used as it excludes any interference from the virtualization system. The resulting dataset contains 233 system variables monitored at runtime and focuses on two failure modes: \textit{System Hang} and \textit{Crash}. 
		
		
		
		\subsubsection{Methodology}
		For this case study, only \textit{Variance} and \textit{Correlation} dimensionality techniques were used, alongside a Z-score normalization. The algorithms considered were: \textit{\gls{SVM}, \gls{DT}, \gls{RF}, Bagging, Extra Trees, \gls{NN},} and \textit{\gls{KNN}}. The initial configurations of the different methods were the same as for the previous case study. To briefly analyze the effect of the imbalance in the data, a simple undersampling method, \textit{Random Undersampling}, was also included. Finally, three configurations were tested for the pair $\Delta t_l, \Delta t_p$: [20,20], [40,20], and [60,20], considering a "short", "medium", and "long" term prediction. The samples were labeled according to the approach proposed in \cite{Salfner2010}. Samples for which  no failures were observed within the chosen interval will be referred to as \textit{Control}. As an example, the classes distribution for $\Delta t_l = 40, \Delta t_p = 20 $ is 48937, 832, 145, for Control, Hang, and Crash respectively. Given the number of combinations, only those deemed relevant will be detailed. 
		
		%
		
		Due to the scope of this case study (and time-constraints) no fine-tuning was conducted. The performance estimation was calculated using 10-fold cross-validation. As failures are rare events, most of the collected data pertains to non-failure prone systems, and thus the dataset was limited to 50000 samples. 
		
		\subsubsection{Results}
		Due to space constraints, the following results pertain only to the values of $\Delta t_l = 40, \Delta t_p = 20$, as this was the configuration that achieved best results. As in the previous study, an initial analysis was conducted to better understand the problem. However, this is both a multi-class and a time-series problem, which requires a different approach. \propheticus also handles this type of problem and contains techniques specifically for sequential data (e.g. \textit{sliding window}).

			%

		As our goal in this description of the case study is to complement the previous one, the analysis of the dataset will be omitted. Concerning the performance of the models, even with no preprocessing techniques, some algorithms were able to perform well (e.g. \gls{DT}) while others were not (e.g. \gls{SVM}), as can be seen in \textit{Table \ref{tbl:summary_results}}. The \gls{DT} algorithm achieved the best $F_2$-score and informedness, making it the best choice for the most critical scenarios (i.e. business- and heightened-critical), followed by Bagging and Extra Trees. Using feature selection slightly improved the performance of some algorithms, and undersampling allowed some to predict even more failures, although this came at the expense of significantly more \glspl{FP}.
		
		\begin{table}[!tb]
			\caption{Multi-Class Summary Real Dataset Results}
			\centerline{\includegraphics[width=0.48\textwidth]{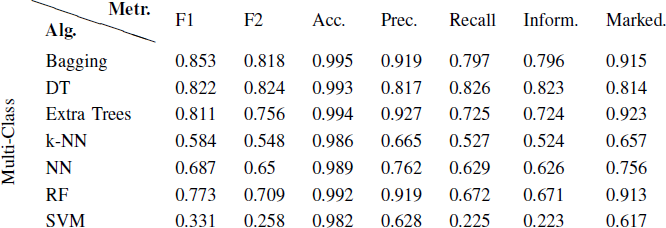}}
			\label{tbl:summary_results}
		\end{table}
		
		When applying techniques like dimensionality reduction and data sampling, which usually incur in some performance trade-off, the choice of the best model becomes far more complicated. Hence, \propheticus was used to rank the performance of the best algorithms (\gls{DT}, Bagging, Extra Trees, and \gls{RF}) along with feature selection and undersampling techniques for the various scenarios. As can be observed in \textit{Table \ref{tbl:rank_multiclass_all}}, different combinations generate the best models for each scenario. In fact, although the use of undersampling produced models with low precision (due to the high number of \glspl{FP}), such models can be of interest for scenarios that are more concerned with correctly predicting both the positive and negative samples (e.g. Heightened-critical). Moreover, the best model for each scenario is obtained through the use of feature selection.	
		\begin{table}[!tb]
			\caption{Multi-class Complete Rank per Scenario}
			\centerline{\includegraphics[width=0.34\textwidth]{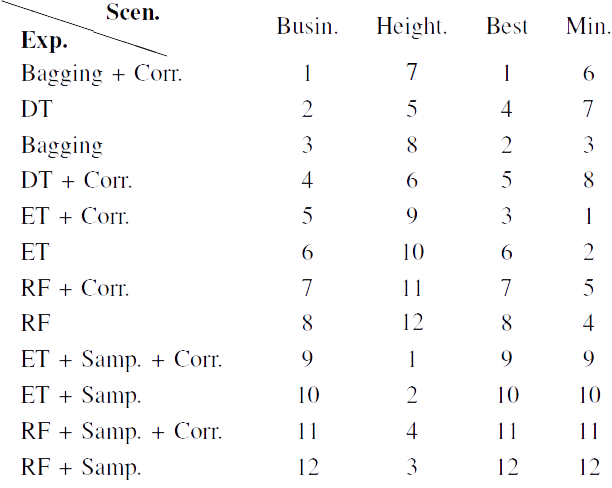}}
			\label{tbl:rank_multiclass_all}
		\end{table}
		
		Sound research requires statistical comparisons to be done to state that a given algorithm is better than the alternative. The base \gls{DT} and Bagging algorithms were chosen as they were the best for the most demanding scenarios. The comparison was done using the $F2$-score results of each run, stored in the generated logs. As the dataset was the same for both experiments, a paired statistical test must be used. To decide between parametric and non-parametric tests, the normality of the data (using the \textit{Lilliefors} and \textit{Shapiro-Wilk} test) and the homogeneity of variance (using the \textit{Levene} test) were analyzed. Both conditions were satisfied, so the \textit{T-Test} parametric test was chosen \cite{Field2013}, which gave a $p\_value = 0.0215$. With this $p\_value$, it is safe to state that the two algorithms are in fact different for a significance level of 5\% ($p\_value < 0.05$). However, when comparing Bagging with feature selection by correlation and \gls{DT} there were no significant differences.
		
		\subsubsection{Discussion}
		Various algorithms and techniques achieved good results, although the best were mostly based on the \gls{DT} algorithm. As expected, undersampling the majority class improved the ability of some algorithms to predict failures, yet, this came at the expense of a significant number of \glspl{FP}. Still, the best algorithms achieved very good performance, with \gls{DT} reporting 82\% and 83\% of precision and recall, respectively.
		
		This case study demonstrated that \propheticus can be used for a significantly different problem and data. Besides having functionalities that take into account the sequence (time dimension), it also has the ability to easily compare and rank different experiments according to the preferences of the user. Finally, \propheticus also provides the functionalities to statistically compare results, essential for any sound research. 
		
		If we were using only Scikit-learn (and not \propheticus), the comparison and ranking of the different experiments would have to be done manually and considering the various metrics. To statistically compare the experiments, it would be necessary to use specific packages and implement the code required to validate the different assumptions and call the adequate tests.

		\section{Demo}
		\label{sec:demo}
		This section briefly demonstrates how \propheticus can be used to assist in the development of predictive models for reliable and secure software, partially following the vulnerabilities case study presented in \textit{Section \ref{sec:case_studies}}. Due to space restrictions, only the most relevant parts will be shown. Notwithstanding, a comprehensive video demo is also available\footnotemark[\getrefnumber{footnote:demo}].
		
\begin{figure}
			\begin{minipage}[c]{0.48\linewidth}
				\vspace*{2mm}
				\begin{center}
					\includegraphics[width=1\textwidth]{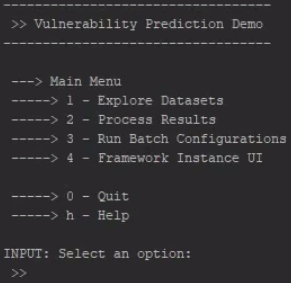}
				\end{center}
				\vspace*{2mm}
				\caption{Initial Menu}
				\label{fig:demo:initial_menu}
			\end{minipage}
			\hspace{0.02\linewidth} 
			\begin{minipage}[c]{0.48\linewidth}
				\begin{center}
					\includegraphics[width=1\textwidth]{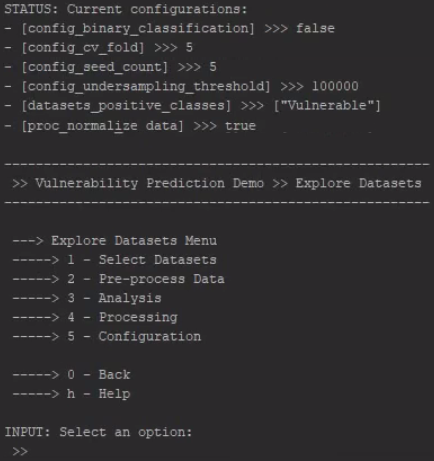}
				\end{center}
				\caption{Explore Datasets Menu}
				\label{fig:demo:explore_datasets}
			\end{minipage}
		\end{figure}

		The initial menu of \propheticus (depicted in \textit{Fig. \ref{fig:demo:initial_menu}}) gives access to its high-level functionalities. Briefly, option \textit{1)} allows the user to explore the dataset and create predictive models; \textit{2)} allows processing the results (e.g. compare/archive); \textit{3)} executes the experiments defined for batch processing (which allows to quickly set up a comprehensive experimental procedure, combining multiple techniques and parameters); \textit{4)} provides a \textit{hook} for a custom (user-defined) menu (e.g. to generate the datasets); and finally, options \textit{0)} and \textit{h)} exit the tool and display the help menu, respectively. Another feature that is always visible is a \textit{navigation path}, which quickly indicates the current position within the tool.
		
		When dealing with a new dataset, the first step is to explore it and test different algorithms and techniques. After choosing option \textit{1)} new information and menus are provided, as shown in \textit{Fig. \ref{fig:demo:explore_datasets}}. First of all, the description of the currently defined configurations is presented at the top, followed by the navigation path (which now reflects the chosen menu). Both components are always visible throughout the tool, but due to space restrictions, the following images will be truncated to fit only the most relevant content. Briefly, option \textit{1)} allows selecting the datasets; \textit{2)} allows defining which preprocessing techniques (and respective configurations) to use; \textit{3)} provides several tools to analyze the datasets; \textit{4)} allows selecting, configuring, and running different algorithms; and \textit{5)} allows modifying several parameters of \propheticus' (e.g. \# of runs).
		
\begin{figure}
			\begin{minipage}[c]{0.48\linewidth}
				\begin{center}
					\includegraphics[width=1\textwidth]{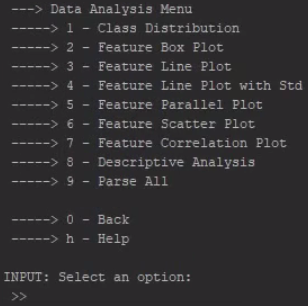}
				\end{center}
				\caption{Analysis Menu}
				\label{fig:demo:analysis}
			\end{minipage}
			\hspace{0.02\linewidth} 
			\begin{minipage}[c]{0.48\linewidth}
				\vspace*{2mm}
				\begin{center}
					\includegraphics[width=1\textwidth]{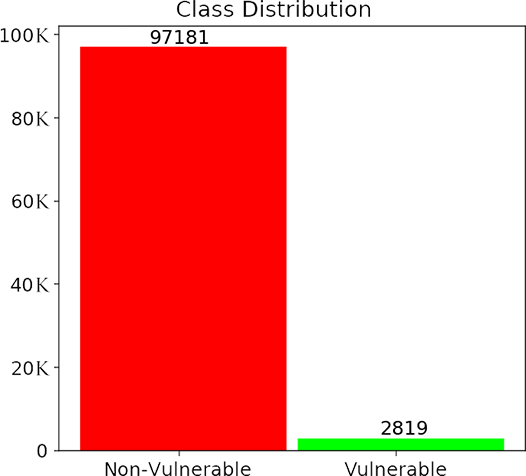}
				\end{center}
				\vspace*{2mm}
				\caption{Class Distribution}
				\label{fig:demo:classdistro}
			\end{minipage}
		\end{figure}

\begin{figure}
			\begin{minipage}[c]{0.48\linewidth}
				\vspace*{5mm}
				\begin{center}
					\includegraphics[width=1\textwidth]{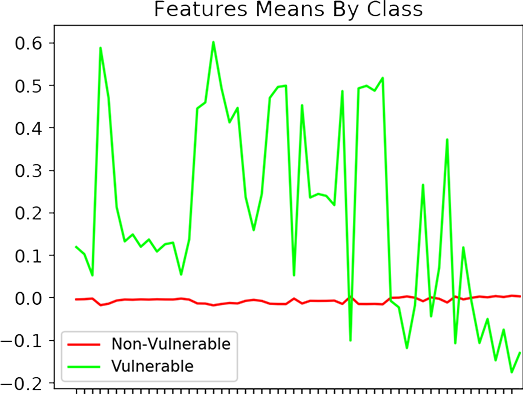}
				\end{center}
				\vspace*{6mm}
				\caption{Features Means}
				\label{fig:demo:lineplot}
			\end{minipage}
			\hspace{0.02\linewidth} 
			\begin{minipage}[c]{0.48\linewidth}
				\begin{center}
					\includegraphics[width=1\textwidth]{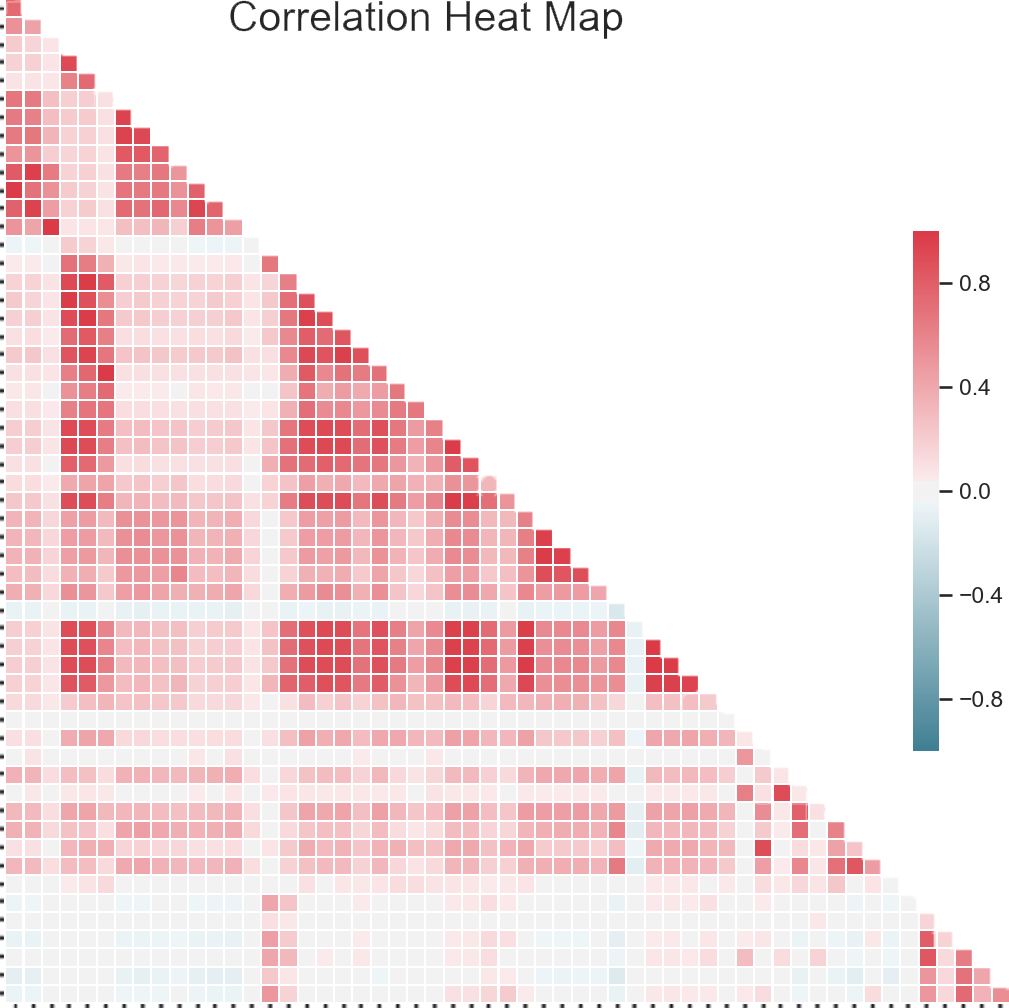}
				\end{center}
				\caption{Features Correlation}
				\label{fig:demo:heatmap}
			\end{minipage}
		\end{figure}

		After selecting the intended datasets (through option \textit{1}), the next step is to analyze it,  using option \textit{3)} (option 2 will be used later on when exploring preprocessing techniques). Several options are available (depicted in \textit{Fig. \ref{fig:demo:analysis}}), such as: \textit{1)} visualizing the distribution of the samples per class; \textit{2)} analyzing the distribution of the values per feature; \textit{3} and \textit{4)} plotting the average values of each class per feature; \textit{5)} plotting a parallel coordinates plot; \textit{6} and \textit{7)} analyzing the correlation of features; and \textit{8)} generating a report using descriptive analysis.
		Referring to the vulnerability case study, using the class distribution analysis (option \textit{1}) it was possible to observe that the datasets were highly imbalanced (as can be seen in \textit{Fig. \ref{fig:demo:classdistro}}), which influences which techniques and metrics should be used. Additionally, by observing the averages of the values of the features per class (option \textit{3}, depicted in \textit{Fig. \ref{fig:demo:lineplot}}) there appear to be noticeable differences between the classes, suggesting that they may be distinguishable. Finally, by using a heat map (option \textit{7}, depicted in \textit{Fig. \ref{fig:demo:heatmap}}) it was possible to observe that there are highly correlated features, which likely can be removed without losing too much information. 
		
\begin{figure}
			\begin{minipage}[c]{0.48\linewidth}
				\vspace*{4mm}
				\begin{center}
					\includegraphics[width=1\textwidth]{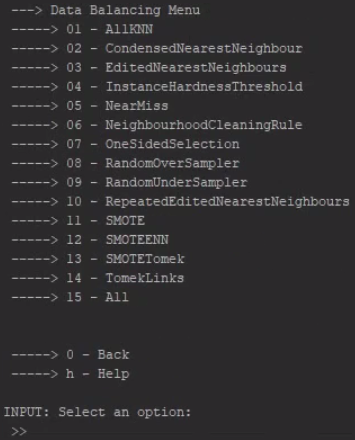}
				\end{center}
				\vspace*{5mm}
				\caption{Sampling Techniques}
				\label{fig:demo:sampling}
			\end{minipage}
			\hspace{0.02\linewidth} 
			\begin{minipage}[c]{0.48\linewidth}
				\begin{center}
					\includegraphics[width=1\textwidth]{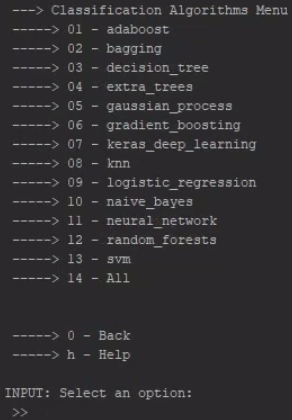}
				\end{center}
				\caption{Classification Algorithms}
				\label{fig:demo:classalgorithms}
			\end{minipage}
		\end{figure}

		The next step consists of assessing how the algorithms perform using the complete dataset, which is done by returning to the \textit{Explore Datasets} (\textit{Fig. \ref{fig:demo:explore_datasets}}) menu and selecting option \textit{4)}. For simplicity (and space restrictions) this demo focuses only on classification tasks (which can be seen in \textit{Fig. \ref{fig:demo:classalgorithms}}) and uses only the \gls{RF} algorithm (option \textit{12}).
		%
		After completing the experiments, \propheticus generates several reports to assist in the interpretation of the results (as detailed in \textit{Section \ref{sec:framework:modules}}). By analyzing the \gls{RF} results, it was possible to observe that they were similar to those obtained using the \gls{DT} algorithm (\textit{Fig. \ref{fig:step1_dt_cm}}). Thus, different preprocessing techniques should be used to attempt to deal with the characteristics of the dataset.
		
\begin{figure}
			\begin{minipage}[c]{0.48\linewidth}
				\begin{center}
					\includegraphics[width=1\textwidth]{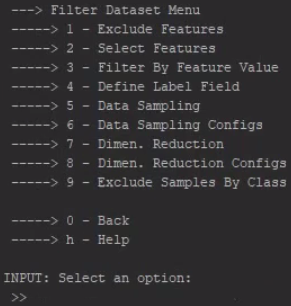}
				\end{center}
				\caption{Preprocess Menu}
				\label{fig:demo:preprocess}
			\end{minipage}
			\hspace{0.02\linewidth} 
			\begin{minipage}[c]{0.48\linewidth}
				\vspace*{0.5mm}
				\begin{center}
					\includegraphics[width=1\textwidth]{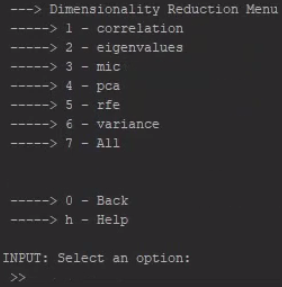}
				\end{center}
				\vspace*{0.5mm}
				\caption{Dimensionality Reduction}
				\label{fig:demo:dimred}
			\end{minipage}
		\end{figure}

		Returning to the \textit{Explore Datasets} menu (\textit{Fig. \ref{fig:demo:explore_datasets}}) and selecting option \textit{2)} provides a new menu (depicted in \textit{Fig. \ref{fig:demo:preprocess}}) that allows choosing and configuring techniques such as sampling (options \textit{5} and \textit{6}) and dimensionality reduction (options \textit{7} and \textit{8}). 
		Other techniques often needed within dependability research are also available, for example: \textit{1} and \textit{2)} excluding/include specific features; \textit{3)} filtering the samples based on specific features values; and \textit{9)} excluding samples by a specific class.
		Based on the previous analyses, highly correlated features and those with null variance were removed. Additionally, the undersampling technique \textit{InstanceHardnessThreshold} (from Imbalanced-learn \cite{imblearn}) was used to balance the datasets. The list of techniques available for dimensionality reduction and sampling can be seen in \textit{Fig. \ref{fig:demo:dimred}} and \textit{\ref{fig:demo:sampling}} respectively. 
		Running the \gls{RF} algorithm with these configurations considerably improved the results, having performance similar to the one depicted in \textit{Fig. \ref{fig:step5_rf_cm}}. Finally, combining these techniques with fine-tuning (enabled through the \textit{Configuration} menu) will reach the exact same results obtained in the case study.
		
		After the experiments are finished, it is necessary to compare the results. \propheticus provides a functionality that allows choosing which experiments to compare (as described in \textit{Section \ref{sec:framework:modules}}), which then generates a comparison report, including some complementary graphics (e.g. barplots for performance/time metrics) and statistical comparisons (e.g. normality, homogeneity, and statistical/hypothesis testing).

		Although \propheticus already comprises several methods, the state-of-the-art evolves continuously. For researchers, the use of such advances is often promising, and thus, \propheticus was developed taking into account the need to expand. This concern is something that distinguishes it from alternative \gls{ML} frameworks (e.g. using Weka, (if a package is not available) it is necessary to extract its source code, add the algorithm and modify some required files, and recompile Weka; other tools are not even meant to be extended). To achieve this, \propheticus implements a plugin paradigm, where each technique's interface follows a standard (identical to Scikit-learn) and merely needs to be declared. As an example, to add a new algorithm one only needs to add a small Python dictionary (similar to \textit{Listing \ref{lst:algorithm_code}}) to the intended structure (e.g. algorithms) indicating the package where it is available (which can be embedded in \propheticus or an external package, e.g. Keras), its callable, and its arguments (which can then also be used to define values through the CLI). Moreover, this paradigm is expanded to other sections, such as sampling/dimensionality reduction techniques and even performance metrics.

		\lstset{%
			basicstyle=\ttfamily\footnotesize, 
			breaklines = true,
			tabsize=2,
			showstringspaces=false,
			upquote=true
		}
		\begin{lstlisting}[language=Python,caption={Algorithm Declaration},label={lst:algorithm_code}]
			'decision_tree':{
				'package':'sklearn.tree',
				'callable':'DecisionTreeClassifier',
				'parameters':{
					'criterion':{'type':'str','values':['gini','entropy']},
					'max_depth':{'type':['int','None']},
					...
			}}
		\end{lstlisting}	

		\section{Discussion and Conclusion}
		\label{sec:conclusion}

		Software complexity is growing significantly and traditional validation techniques are not able to keep up, leading to the use of \gls{ML} to assist them. However, due to a limited offer of existing tools, most researchers are required to set up their own experimental \gls{ML} workflow. Besides consuming time and effort, this also increases the chance of faulty code that can compromise the results. \propheticus is a generalizable \gls{ML} framework designed by researchers for researchers that attempts to overcome this and facilitate the use of \gls{ML} for research while minimizing the need for coding. By creating predictive models it can then be used to assist traditional techniques in developing more dependable systems.

		Although there are already several tools for \gls{ML} each has certain characteristics that either render them less appropriate for research or require considerable \gls{ML} knowledge. Due to space restrictions, it is not possible to enumerate all the limitations/advantages of each framework and how \propheticus relates/improves them. Still, we believe that the arguments presented in \textit{Section \ref{sec:related}}, the case studies in \textit{Section \ref{sec:case_studies}}, and the short demo in \textit{Section \ref{sec:demo}} provide enough details to demonstrate that \propheticus is different from the others. \propheticus is not a competitor with existing solutions, but an alternative flexible framework focused on research. It attempts to overcome the existing limitations of general-purpose frameworks (e.g. Weka) that drive users to use more low-level libraries (e.g. Scikit-learn) which require far more \gls{ML} expertise.
		
		\propheticus is a tool to facilitate/assure the correct experimental use of ML by dependability researchers. It was developed considering the tasks/challenges they often face (e.g. imbalanced datasets, sensitive features, accidental \enquote{false-predictors}). Besides influencing which techniques were initially included this also guided its design and workflow. 
		
		
		\propheticus is already a comprehensive framework, and thus it is not possible to include all the features in the demo presented in this paper. Notwithstanding, a more comprehensive video demo is available \footnotemark[\getrefnumber{footnote:demo}]. Regarding future work, the main focus will be on broadening its scope and functionalities.
		


		
		\def\IEEEbibitemsep{0pt plus .5pt}
		\bibliographystyle{IEEEtran}
		\bibliography{IEEEabrv,bibliography}


	\end{document}

%% file: glossary.tex
\newacronym{AI}{AI}{Artificial Intelligence}
\newacronym{API}{API}{Application Programming Interface}
\newacronym{AUC}{AUC}{Area Under the Curve}
\newacronym{ADASYN}{ADASYN}{Adaptive Synthetic}

\newacronym{CART}{CART}{Classification and Regression Trees}
\newacronym{CLI}{CLI}{Command Line Interface}
\newacronym{CM}{CM}{Confusion Matrix}

\newacronym{DT}{DT}{Decision Tree}

\newacronym{EC}{EC}{Evolutionary Computation}

\newacronym{FIR}{FIR}{Fault Injection Run}

\newacronym{GPU}{GPU}{Graphics Processing Unit}
\newacronym{GR}{GR}{Golden Run}
\newacronym{GSWFIT}{G-SWFIT}{Generic Software Fault Injection Technique}

\newacronym{HPC}{HPC}{High Processing Computing}

\newacronym{KNN}{k-NN}{k Nearest Neighbors}

\newacronym{LBFGS}{LBFGS}{Limited-memory Broyden–Fletcher–Goldfarb–Shanno}

\newacronym{ML}{ML}{Machine Learning}

\newacronym{NN}{NN}{Neural Network}

\newacronym{OFP}{OFP}{Online Failure Prediction}
\newacronym{OS}{OS}{Operating System}
\newacronym{OVO}{OVO}{One-Versus-One}
\newacronym{OVR}{OVR}{One-Versus-Rest}

\newacronym{PCA}{PCA}{Principal Component Analysis}

\newacronym{RBF}{RBF}{Radial Basis Function}
\newacronym{ROC}{ROC}{Receiver Operating Characteristics}
\newacronym{RF}{RF}{Random Forest}
\newacronym{RFE}{RFE}{Recursive Feature Elimination}

\newacronym{SWIFI}{SWIFI}{Software Implemented Fault Injection}
\newacronym{SVM}{SVM}{Support Vector Machine}
\newacronym{SMOTE}{SMOTE}{Synthetic Minority Over-sampling Technique}

\newacronym{FP}{FP}{False Positive}
\newacronym{FN}{FN}{False Negative}
\newacronym{TN}{TN}{True Negative}
\newacronym{TP}{TP}{True Positive}